\documentclass[10pt,twocolumn,letterpaper]{article}
\usepackage[pagenumbers]{cvpr} % To force page numbers, e.g. for an arXiv version
\usepackage{float}
\usepackage{stfloats}

\definecolor{cvprblue}{rgb}{0.21,0.49,0.74}
\usepackage[pagebackref,breaklinks,colorlinks,allcolors=cvprblue]{hyperref}
\title{On the Limitations of Vision-Language Models in Understanding Image Transforms}

\author{Ahmad Mustafa Anis$^{1}$, Hasnain Ali$^{2}$, M. Saquib Sarfraz$^{3}$\\[1em]
$^{1}$Cohere for AI Community, $^{2}$Arbisoft, $^{3}$Karlsruhe Institute of Technology\\
{\tt\small Corresponding to: ahmadanis5050@gmail.com}
}

\begin{document}
\maketitle
\begin{abstract}
Vision Language Models (VLMs) have demonstrated significant potential in various downstream tasks, including Image/Video Generation, Visual Question Answering, Multimodal Chatbots, and Video Understanding. However, these models often struggle with basic image transformations. This paper investigates the image-level understanding of VLMs, specifically CLIP by OpenAI and SigLIP by Google. Our findings reveal that these models lack comprehension of multiple image-level augmentations. To facilitate this study, we created an augmented version of the Flickr8k dataset, pairing each image with a detailed description of the applied transformation. We further explore how this deficiency impacts downstream tasks, particularly in image editing, and evaluate the performance of state-of-the-art Image2Image models on simple transformations.

\end{abstract}

\section{Introduction}
Vision Language Models like CLIP~\cite{radford2021learningtransferablevisualmodels} and SigLIP~\cite{zhai2023sigmoidlosslanguageimage} have emerged as powerful frameworks that incorporate visual and text encoders aligned via large-scale pre-training on image-text pairs. These models have demonstrated impressive performance across various downstream tasks, including Text-to-Image Generation~\cite{rombach2022highresolutionimagesynthesislatent, ramesh2021zeroshottexttoimagegeneration}, Video Action Recognition~\cite{wang2021actionclipnewparadigmvideo}, and applications in the Biomedical domain~\cite{zhang2024biomedclipmultimodalbiomedicalfoundation}. CLIP-like pre-training has been extended to other modalities as well, such as CLAP for Audio and Language~\cite{elizalde2022claplearningaudioconcepts}.

\begin{figure}[t]
    \centering
    \includegraphics[width=\linewidth]{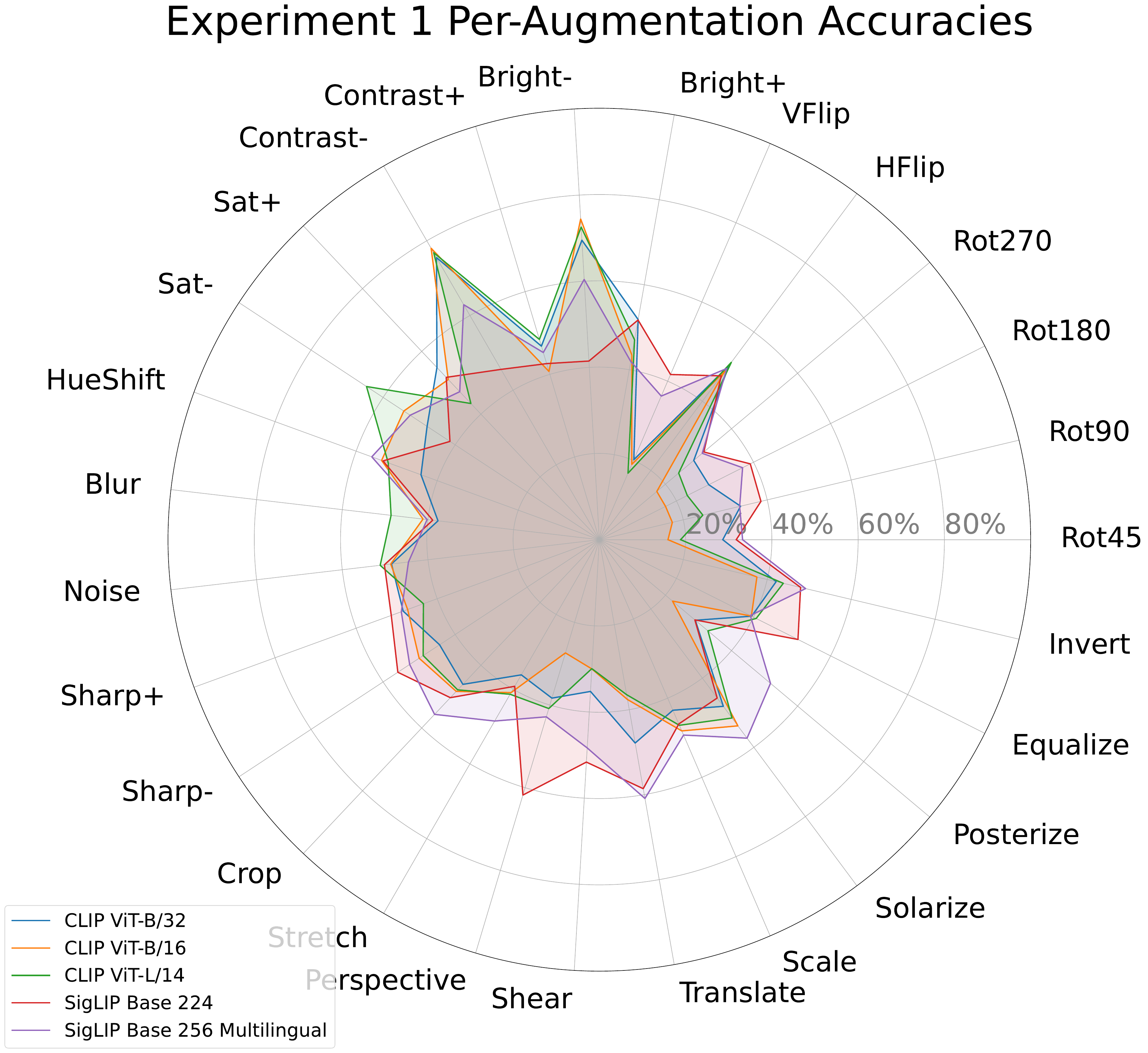}
    \caption{Comparison of image augmentation understanding between humans and Vision Language Models (CLIP/SigLIP). While humans can recognize and describe image transformations like rotation, brightness adjustment, and contrast changes, Vision Language Models show significant limitations in comprehending these basic image manipulations.}
    \label{fig:augmentation_comparison}
\end{figure}

However, despite their broad success, a fundamental question remains unanswered: \textit{``Can Vision Language Embedding Models understand simple Image Transformations?''} This question is particularly crucial as these models are increasingly deployed for image editing tasks, where understanding basic transformations is essential for meaningful manipulation. As shown in Figure~\ref{fig:augmentation_comparison}, our systematic evaluation reveals a significant gap between human and machine understanding of common image modifications.

Understanding image augmentations is fundamental for robust visual reasoning, as real-world images frequently appear with variations in brightness, contrast, rotation, and other transformations. While these models are designed to exhibit robustness and invariant behavior to standard image transforms, we argue that this invariance might come at the cost of explicit understanding. Although invariance was valuable for earlier models trained in data-constrained environments~\cite{he2015deepresiduallearningimage} on datasets like ImageNet~\cite{russakovsky2015imagenetlargescalevisual}, modern foundation models trained on vast amounts of data should ideally possess both invariance when needed and explicit understanding of transformations when required.

Through comprehensive evaluation of CLIP and SigLIP responses to various controlled augmentations, we demonstrate significant limitations in these models' ability to reason about simple image transformations. Our findings have important implications for downstream tasks that rely on these models, particularly in applications requiring explicit understanding of image modifications. This work not only highlights a critical gap in current Vision Language Models but also emphasizes the need for developing approaches that can better capture fundamental aspects of visual reasoning.
\section{Related Works}
\noindent\textbf{Spatial Reasoning:} Multiple works have been done to evaluate spatial reasoning in CLIP-related models. The paper "Visual-Spatial Reasoning"~\cite{liu2023visualspatialreasoning} shows that Vision Language models like CLIP are not good in spatial reasoning. ReCLIP~\cite{subramanian2022reclipstrongzeroshotbaseline} re-purposes CLIP to extend it to tasks related to Spatial Reasoning by introducing a Spatial Relation Resolver. Lewis et al.~\cite{lewis2024doesclipbindconcepts} show that CLIP models perform poorly on compositional visual reasoning tasks and cannot encode compositional concepts or bind variables in a structure-sensitive way (e.g., differentiating ``cube behind sphere'' from ``sphere behind cube''). OmniCLIP~\cite{liu2024omniclipadaptingclipvideo} shows that CLIP falls short in capturing and integrating spatial-temporal features which is essential for video recognition and proposes a framework to extend CLIP for spatial temporal features for video recognition. \\

\noindent\textbf{Linguistic Reasoning:} Studies have shown that CLIP also does not perform well on pure linguistic tasks. Sam et al.~\cite{sam2024finetuningclipreasonpairwise} show that CLIP's embedding space lacks the structure of their purely text-based alternatives (e.g., Text($\textit{``King''}$) $-$ Text($\textit{``Man''}$) $+$ Text($\textit{``Woman''}$) $\approx$ Text($\textit{``Queen''}$)). CyCLIP~\cite{goel2022cyclipcycliccontrastivelanguageimage} demonstrates that image and text representations learned by CLIP are not interchangeable and can lead to inconsistent downstream predictions. \\

\noindent\textbf{Counting:} Counting is an interesting challenge where the model must count the number of entities in an image. Paiss et al.~\cite{paiss2023teachingclipcount} introduce a novel training framework and benchmark to improve the quantitative understanding of VLMs. Ma et al.~\cite{ma2024clipebcclipcountaccurately} enhance CLIP's ability to count with a focus on estimating crowd sizes from images. Zhang et al.~\cite{zhang2024clipcountstarsempirical} studied the question ``Can CLIP Count Stars?'' and showed that CLIP is not reliable in counting stars and contains a quantity bias. \\

\noindent\textbf{Robustness:} Multiple studies have been done to evaluate the robustness of Vision Language Models like CLIP. Tu et al.~\cite{tu2024holisticevaluationrobustnessclip} show that CLIP exhibits strong shape bias. Schlarmann et al.~\cite{schlarmann2024robustclipunsupervisedadversarial} propose an unsupervised adversarial fine-tuning technique to train a robust CLIP vision encoder that is safe against adversarial attacks. Laroudie et al.~\cite{laroudie2023improvingcliprobustnessknowledge} demonstrate that CLIP is \textbf{overconfident} in incorrect predictions, making its predictions less reliable. They also show Domain Shift Vulnerability, where there is a significant accuracy drop when domains are shifted. They propose LP-CLIP, a novel Knowledge distillation framework to improve robustness in CLIP models. \\

\noindent\textbf{3D Understanding:} Recent works have explored CLIP's capabilities in understanding and generating 3D content. CLIP-Forge~\cite{sanghi2022clipforgezeroshottexttoshapegeneration} introduces a zero-shot text-to-shape generation method that addresses the scarcity of paired text-shape data using CLIP's pre-trained image-text representations. Sbrolli et al.~\cite{sbrolli2024captionsproblemcaptionless3dclip} propose unsupervised methods to enhance contrastive text-image-3D alignment by leveraging CLIP's knowledge of textual and 2D data for computing neural perceived similarity between 3D samples. CLIP2Scene~\cite{chen2023clip2scenelabelefficient3dscene} makes the first attempt to transfer CLIP knowledge to 3D scene understanding, achieving impressive results in annotation-free 3D semantic segmentation and fine-tuning scenarios. CISP~\cite{sbrolli2024shapeinfusedjointembeddingsimprove} introduces a framework to enhance 3D shape synthesis from images by aligning 2D images with 3D shapes in a shared embedding space, showing that incorporating explicit 3D knowledge can improve generation coherence compared to standard CLIP-guided models. \\
\section{Dataset \& Augmentation Methodology}
To thoroughly evaluate the image-level understanding of VLMs, we created an augmented version of the Flickr8k~\cite{hodosh2013framing} dataset. This dataset was chosen for its diverse range of images and corresponding captions, providing a robust foundation for our experiments. We developed a systematic approach to apply a variety of image transformations, ensuring each augmented image was paired with a detailed natural language description of the applied modification. This section outlines our data collection process, the specific augmentation techniques employed, and the distribution of these augmentations across the dataset.

\subsection{Data Collection}\label{subsec:datacollection}
We used Flicker8k dataset~\cite{hodosh2013framing} and developed a simple annotation technique to create our augmented dataset. For each image-caption pair, we apply a random augmentation and append the transformation description to the original caption:

\begin{quote}
\small
``A child in a pink dress is climbing up a set of stairs in an entry way, this image has decreased sharpness''
\end{quote}

This approach creates a parallel dataset where each augmented image is paired with an explicitly described transformation.

\subsection{Image Augmentation Methodology}
We implemented 24 image transformations across six categories using PyTorch's \texttt{torchvision.transforms} library\cite{torchvision2016}:

\subsubsection{Geometric Transformations}
\begin{itemize}
    \item \textbf{Rotations:} Four angles (45$^{\circ}$, 90$^{\circ}$, 180$^{\circ}$, 270$^{\circ}$)
    \item \textbf{Flips:} Horizontal and vertical
\end{itemize}

\subsubsection{Color Space Modifications}
Bidirectional adjustments for:
\begin{itemize}
    \item \textbf{Brightness:} $\pm$50\% modifications
    \item \textbf{Contrast:} Similar bidirectional adjustments
    \item \textbf{Saturation:} Controlled adjustments to color intensity
    \item \textbf{Hue:} Warm color shifts (hue=0.1)
\end{itemize}

\subsubsection{Clarity and Focus Transformations}
\begin{itemize}
    \item \textbf{Blur:} Gaussian blur (kernel size 5,5)
    \item \textbf{Sharpness:} Bidirectional modifications ($\pm$50\%)
\end{itemize}

\subsubsection{Geometric Distortions}
\begin{itemize}
    \item \textbf{Perspective:} Controlled shifts (distortion\_scale=0.3)
    \item \textbf{Affine:} Shear (30$^{\circ}$), Translation (20\%), Scale (20\%)
\end{itemize}

\subsubsection{Resolution and Size Modifications}
\begin{itemize}
    \item \textbf{Center Crop:} 224px crop from 256px images
    \item \textbf{Aspect Ratio:} Horizontal stretching (160$\times$256 pixels)
\end{itemize}

\subsubsection{Image Processing Effects}
\begin{itemize}
    \item \textbf{Noise:} Gaussian noise ($\sigma = 0.1$)
    \item \textbf{Intensity:} Solarization (threshold=128), Posterization (2-bit), Equalization
    \item \textbf{Color Inversion:} Complete color space inversion
\end{itemize}

Each augmented image was paired with its original caption plus a description of the applied transformation.

\subsection{Data Distribution}
\begin{figure}[t]
    \centering
    \includegraphics[width=0.95\linewidth]{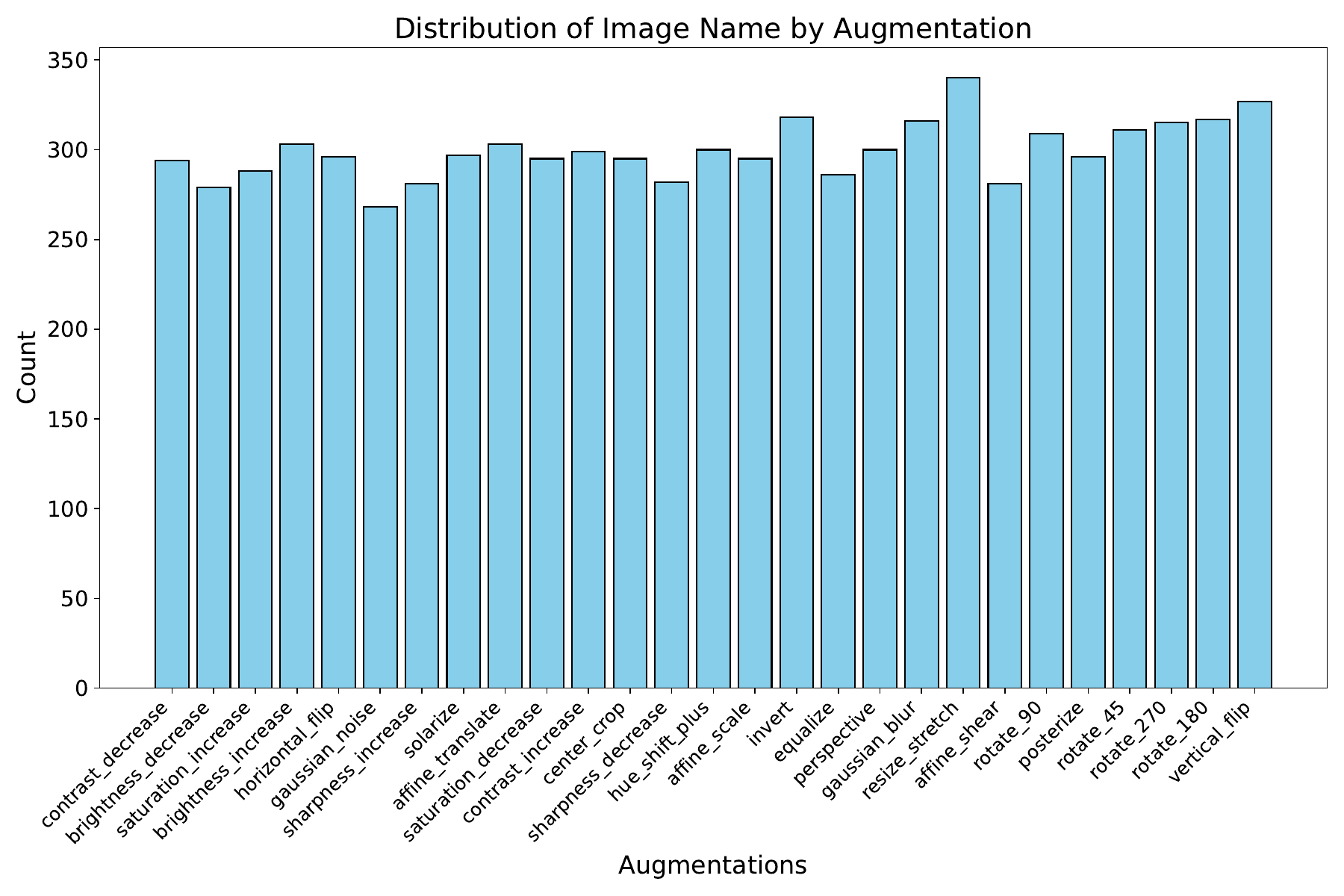}
    \caption{Distribution of individual augmentations applied to the Flickr8k dataset. The augmentations span across multiple transformation types including geometric (rotations, flips), color adjustments (brightness, contrast, saturation), clarity modifications (blur, sharpness), and various image processing effects.}
    \label{fig:augmentation_dist}
\end{figure}

\begin{figure}[t]
    \centering
    \includegraphics[width=0.85\linewidth]{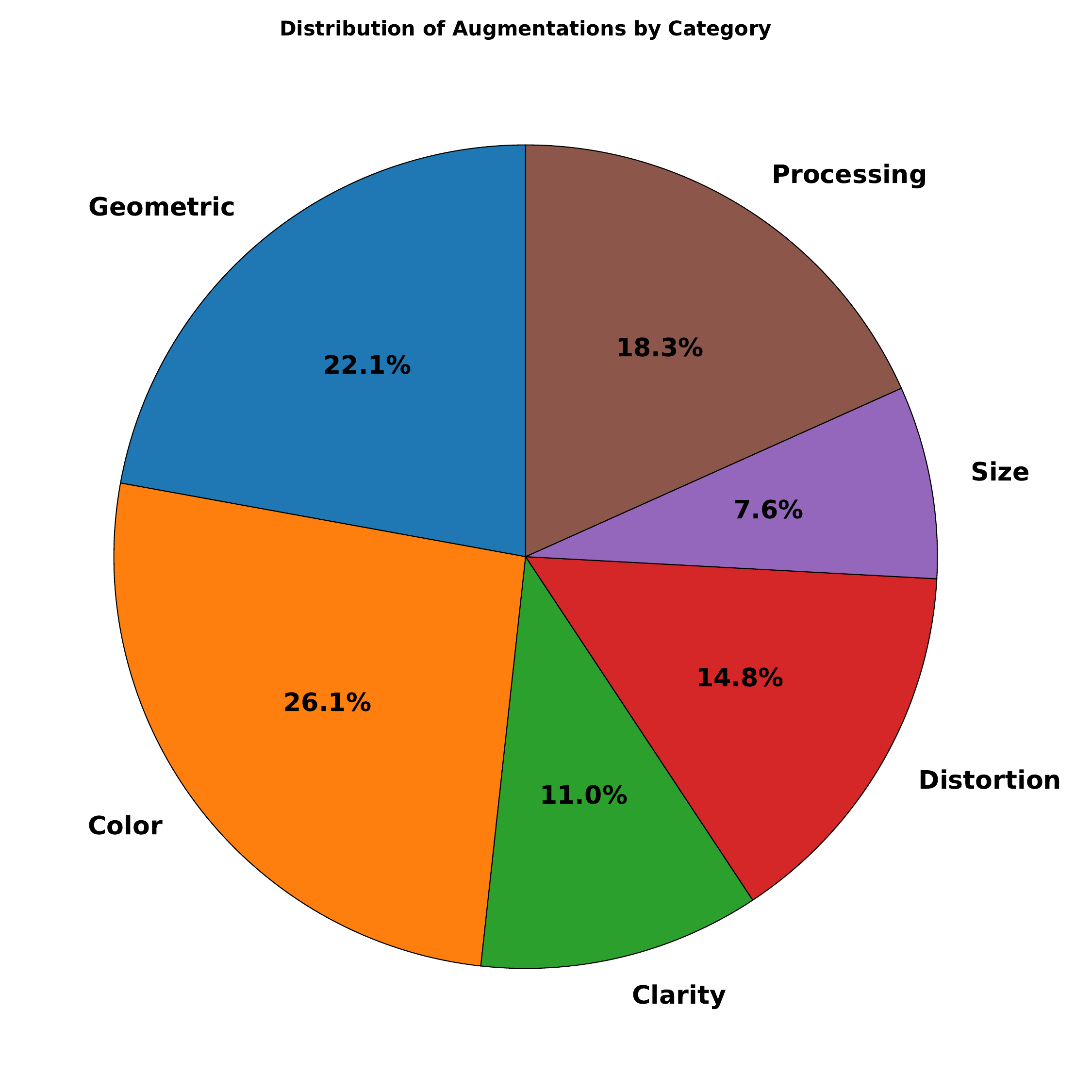}
    \caption{Distribution of augmentations applied to the dataset. The augmentations are grouped into six primary categories: Geometric (rotations and flips), Color (brightness, contrast, saturation, and hue adjustments), Clarity (blur and sharpness), Distortion (perspective and affine transformations), Size (cropping and stretching), and Processing (noise, solarization, posterization, and other effects).}
    \label{fig:aug_categories}
\end{figure}

As shown in Figure~\ref{fig:augmentation_dist}, we implemented diverse augmentations with balanced coverage across transformation types. The categorical analysis (Figure~\ref{fig:aug_categories}) shows that color transformations constitute approximately 43.6\% of all augmentations, followed by processing transformations (41.1\%), distortion effects (7.8\%), and clarity adjustments (7.5\%).

Each category serves a specific evaluation purpose:
\begin{itemize}
    \item \textbf{Geometric:} Tests spatial understanding (rotation, flipping)
    \item \textbf{Color:} Evaluates perception of color variations
    \item \textbf{Clarity:} Assesses recognition under different sharpness levels
    \item \textbf{Distortion:} Tests robustness to perspective and affine changes
    \item \textbf{Size:} Evaluates performance under dimensional changes
    \item \textbf{Processing:} Assesses robustness to image processing effects
\end{itemize}

This distribution ensures comprehensive evaluation of model capabilities across different types of image modifications.
\section{Evaluation of Vision Language Models}
In this section, we present a comprehensive evaluation of VLMs, specifically focusing on their ability to understand and process image augmentations. Our evaluation is structured into three key experiments: understanding augmented descriptions, matching augmented images with descriptions, and classifying image transformations. Each experiment is designed to test different aspects of the models' capabilities, providing a holistic view of their strengths and limitations. Through these experiments, we aim to uncover the extent to which VLMs can accurately interpret and respond to various image modifications, shedding light on their potential and areas for improvement.
\subsection{Understanding Augmented Descriptions}
We first assess the ability of VLMs to accurately associate textual descriptions of image augmentations with their corresponding modified images. This evaluation aims to determine whether the models can comprehend and link the specified transformations described in text to the visual alterations present in the images. By examining the relationship between the augmented descriptions and the visual changes, we can gauge the models' proficiency in understanding and interpreting basic image modifications.

\subsubsection{Methodology}
For each image-caption pair $(I, C)$, we:
\begin{enumerate}
    \item Generate an augmented image $I_{aug}$ using a random transformation $T$
    \item Create an augmented caption $C_{aug}$ by appending the transformation description to the original caption
    \item Compare similarity scores:
    \begin{itemize}
        \item $s_1 = sim(I_{aug}, C_{aug})$: Similarity between augmented image and augmented caption
        \item $s_2 = sim(I_{orig}, C_{aug})$: Similarity between original image and augmented caption
    \end{itemize}
    \item Consider prediction correct if $s_1 > s_2$
\end{enumerate}

Mathematically, the accuracy is computed as:
\begin{equation}
    \text{Accuracy} = \frac{1}{N}\sum_{i=1}^N \mathbb{1}[sim(I_{aug}^{(i)}, C_{aug}^{(i)}) > sim(I_{orig}^{(i)}, C_{aug}^{(i)})]
\end{equation}
where $N$ is the total number of samples and $\mathbb{1}[\cdot]$ is the indicator function.

\subsubsection{Results}
\begin{figure*}[t]
    \centering
    \includegraphics[width=\textwidth]{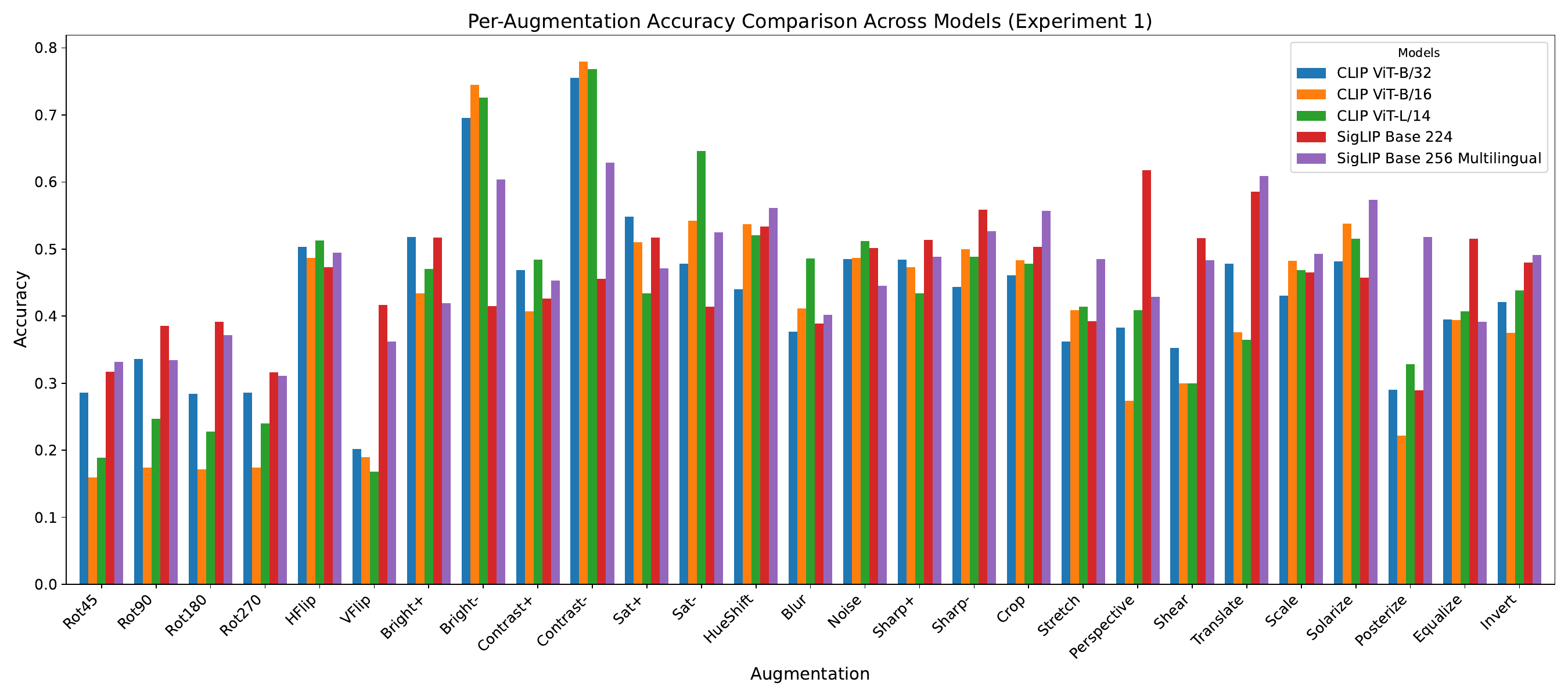}
    \caption{Accuracy comparison of model performance on augmented prompt recognition. Higher values indicate better understanding of the relationship between textual descriptions of transformations and their visual manifestations.}
    \label{fig:exp1_results}
\end{figure*}

\begin{figure}[t]
    \centering
    \includegraphics[width=\columnwidth]{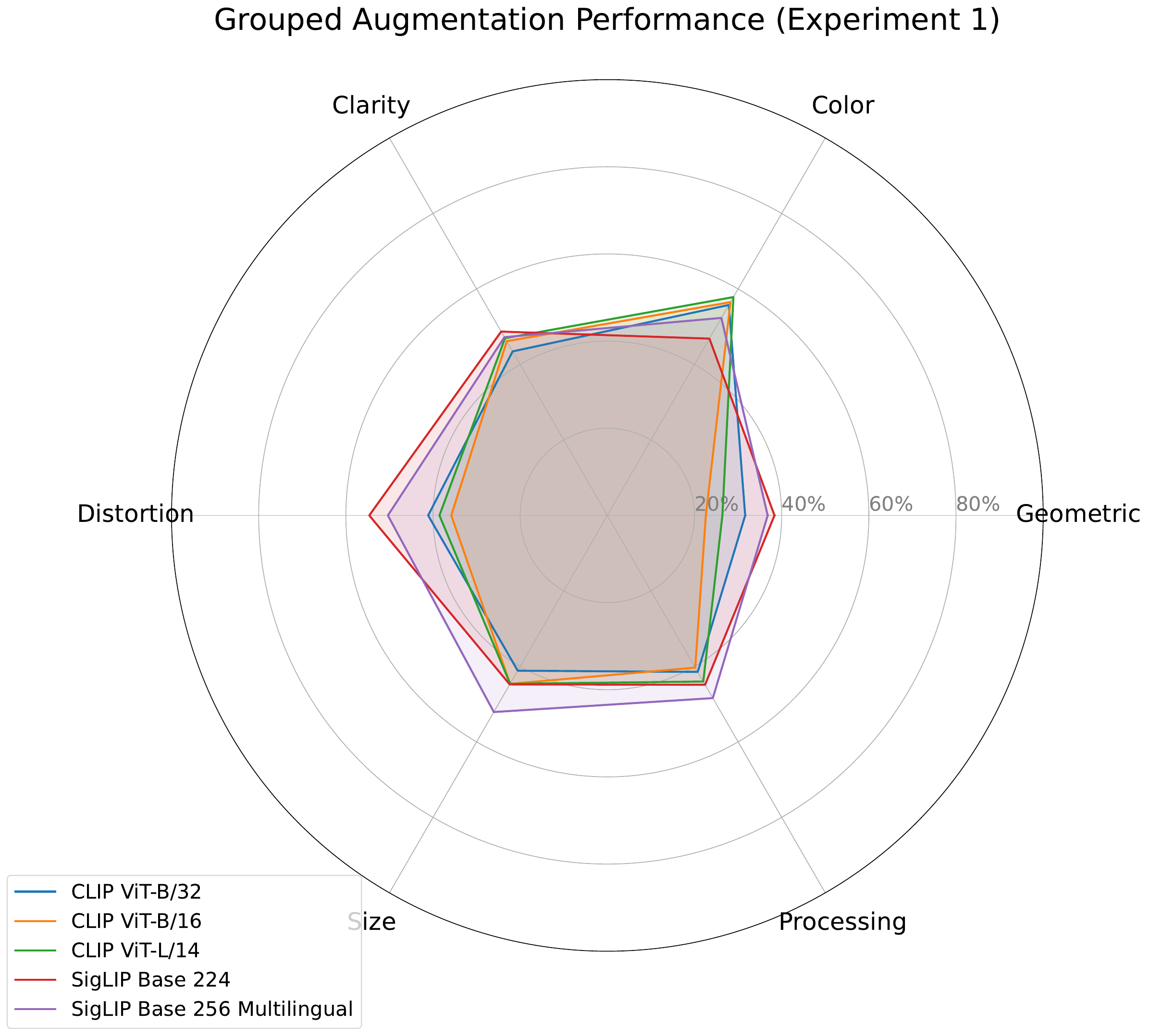}
    \caption{Comparison of model performance on augmentations grouped according to their properties.}
    \label{fig:exp1_results_grouped}
\end{figure}

\begin{table}[htbp]
\centering
\caption{Experiment 1 Overall Accuracy Comparison Across Models}
\begin{tabular}{l c}
\toprule
\textbf{Model} & \textbf{Accuracy (\%)} \\
\midrule
CLIP ViT-B/32 & 42.80 \\
CLIP ViT-B/16 & 40.87 \\
CLIP ViT-L/14 & 43.10 \\
SigLIP Base 224 & 45.78 \\
SigLIP Base 256 Multilingual & 47.21 \\
\bottomrule
\end{tabular}
\label{tab:experiment1_accuracy}
\end{table}

% \subsubsection{Analysis}
Table~\ref{tab:experiment1_accuracy} shows the accuracy across different model variants.

\begin{itemize}
    \item \textbf{Model Architecture Impact:} Larger models (e.g., ViT-L/14) generally show improved performance, suggesting that increased model capacity helps in understanding transformation descriptions. Similarly, CLIP models seem to perform better compared to SigLIP models on some individual types of transformations, as shown in Figure \ref{fig:exp1_results} but SigLIP outperforms CLIP when comparing mean accuracy.
    \item \textbf{Transformation Types:} Models show varying performance across different types of augmentations.
    CLIP and SigLIP perform better in Color and Distortion based augmentations as compared to rest of augmentations however SiGLIP seems to perform better in size and processing based augmentations as shown in Figure \ref{fig:exp1_results_grouped}
    
\end{itemize}

\subsection{Matching Augmented Images with Descriptions}
This evaluation examines the ability of VLMs to accurately match transformed images with their corresponding augmented textual descriptions. The objective is to determine whether the models can effectively identify when an augmented image corresponds to a description that includes specific transformation details, as opposed to a description without such details. By evaluating the models' capacity to link visual modifications with the appropriate textual descriptions, we gain insights into their effectiveness in image-text alignment tasks.

\subsubsection{Methodology}

For each sample \( i \) in the dataset, we perform the following steps:

First, we select an original image \( I^{(i)} \) and apply an augmentation transformation \( T^{(i)} \) to obtain the augmented image:
\begin{equation}
    I_{\text{aug}}^{(i)} = T^{(i)}\left( I^{(i)} \right)
\end{equation}

Next, we prepare the corresponding captions. We obtain the original caption \( C_{\text{orig}}^{(i)} \) associated with \( I^{(i)} \) and define the textual description of the transformation \( T^{(i)} \) as \( \text{desc}\left( T^{(i)} \right) \). The augmented caption is then created by appending the augmentation description to the original caption:
\begin{equation}
    C_{\text{aug}}^{(i)} = C_{\text{orig}}^{(i)} + \text{ ", " } + \text{desc}\left( T^{(i)} \right)
\end{equation}

We compute the similarity between the augmented image and both the original and augmented captions. The similarity with the original caption is:
\begin{equation}
    s_1^{(i)} = \text{sim}\left( I_{\text{aug}}^{(i)},\, C_{\text{orig}}^{(i)} \right)
\end{equation}
and the similarity with the augmented caption is:
\begin{equation}
    s_2^{(i)} = \text{sim}\left( I_{\text{aug}}^{(i)},\, C_{\text{aug}}^{(i)} \right)
\end{equation}
where \( \text{sim}(I, C) \) denotes the similarity function (e.g., cosine similarity) between the embeddings of image \( I \) and caption \( C \).

The model is considered to have correctly associated the augmented image with the augmented caption if:
\begin{equation}
    s_2^{(i)} > s_1^{(i)}
\end{equation}

The overall accuracy over the dataset is computed as:
\begin{equation}
    \text{Accuracy} = \frac{1}{N} \sum_{i=1}^N \mathbb{1}\left[\, s_2^{(i)} > s_1^{(i)} \,\right]
\end{equation}
where \( N \) is the total number of samples, and \( \mathbb{1}[\, \cdot \,] \) is the indicator function defined as:
\begin{equation}
    \mathbb{1}\left[\, \text{condition} \,\right] = \begin{cases}
    1, & \text{if condition is true} \\
    0, & \text{if condition is false}
    \end{cases}
\end{equation}

\subsubsection{Analysis}
\begin{figure*}[t]
    \centering
    \includegraphics[width=0.9\linewidth]{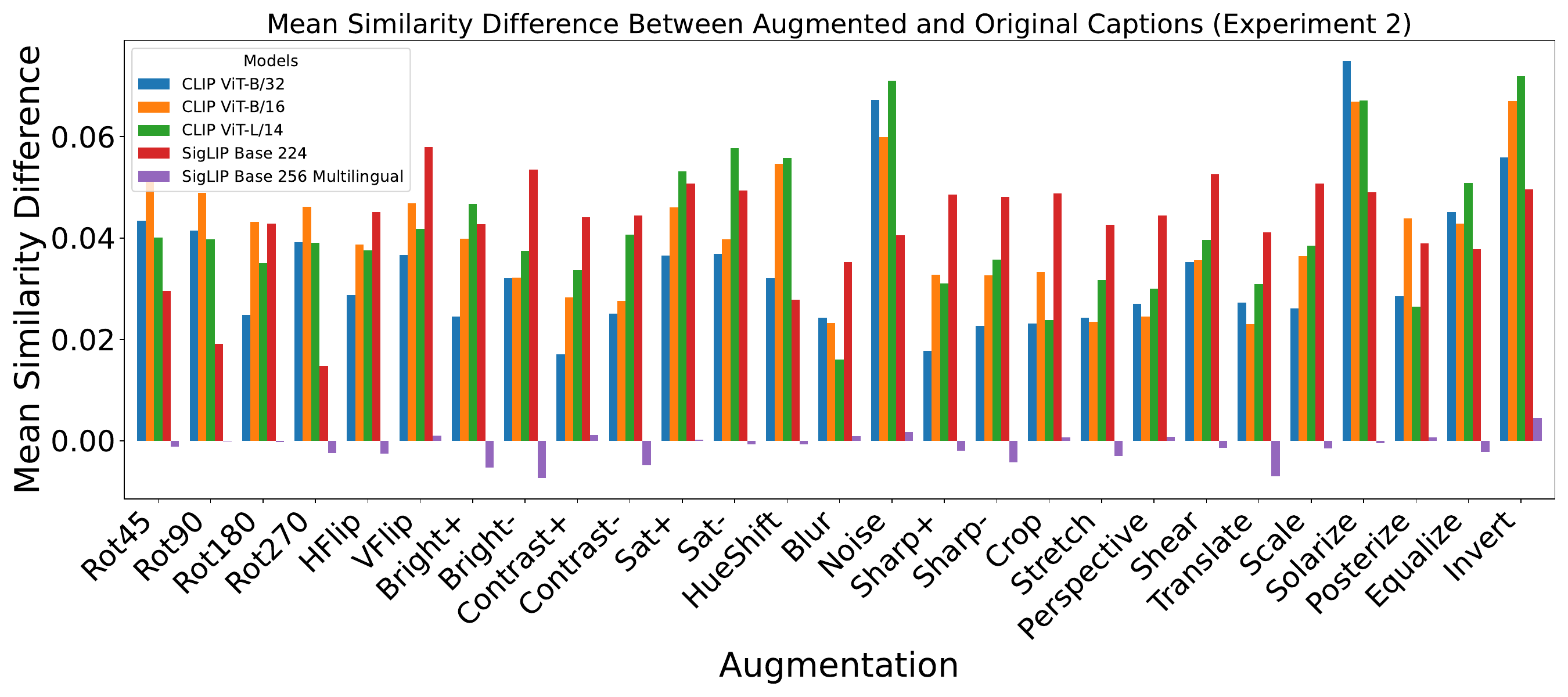}
    \caption{Mean difference between similarity of augmented image with actual prompt and augmented image with augmented prompt }
    \label{fig:exp2_results_mean_diff}
\end{figure*}

The results of Experiment 2 reveal some interesting analysis as shown in Table \ref{tab:experiment2_accuracy}. In experiment 2, all CLIP models perform really well in terms of accuracy showing given an augmented image, Vision Language Models have a better tendency to recognize the augmented prompt in contrast to the actual prompt.
However, \textbf{figure \ref{fig:exp2_results_mean_diff}} shows that there is a very small difference in the similarity score indicating that even though CLIP models perform very well, they can not differentiate between the normal prompt and augmented prompt really well.

\begin{table}[htbp]
\centering
\caption{Experiment 2 Mean Accuracy Comparison}
\begin{tabular}{l c}
\toprule
\textbf{Model} & \textbf{Mean Accuracy} \\
\midrule
CLIP ViT-B/16 & 99.57\% \\
CLIP ViT-B/32 & 98.67\% \\
CLIP ViT-L/14 & 98.15\% \\
SigLIP Base 224 & 64.40\% \\
SigLIP Base 256 Multilingual & 47.41\% \\
\bottomrule
\end{tabular}
\label{tab:experiment2_accuracy}
\end{table}

\subsubsection{Per-Augmentation Analysis}
\begin{figure*}[h]
    \centering
    \includegraphics[width=0.9\linewidth]{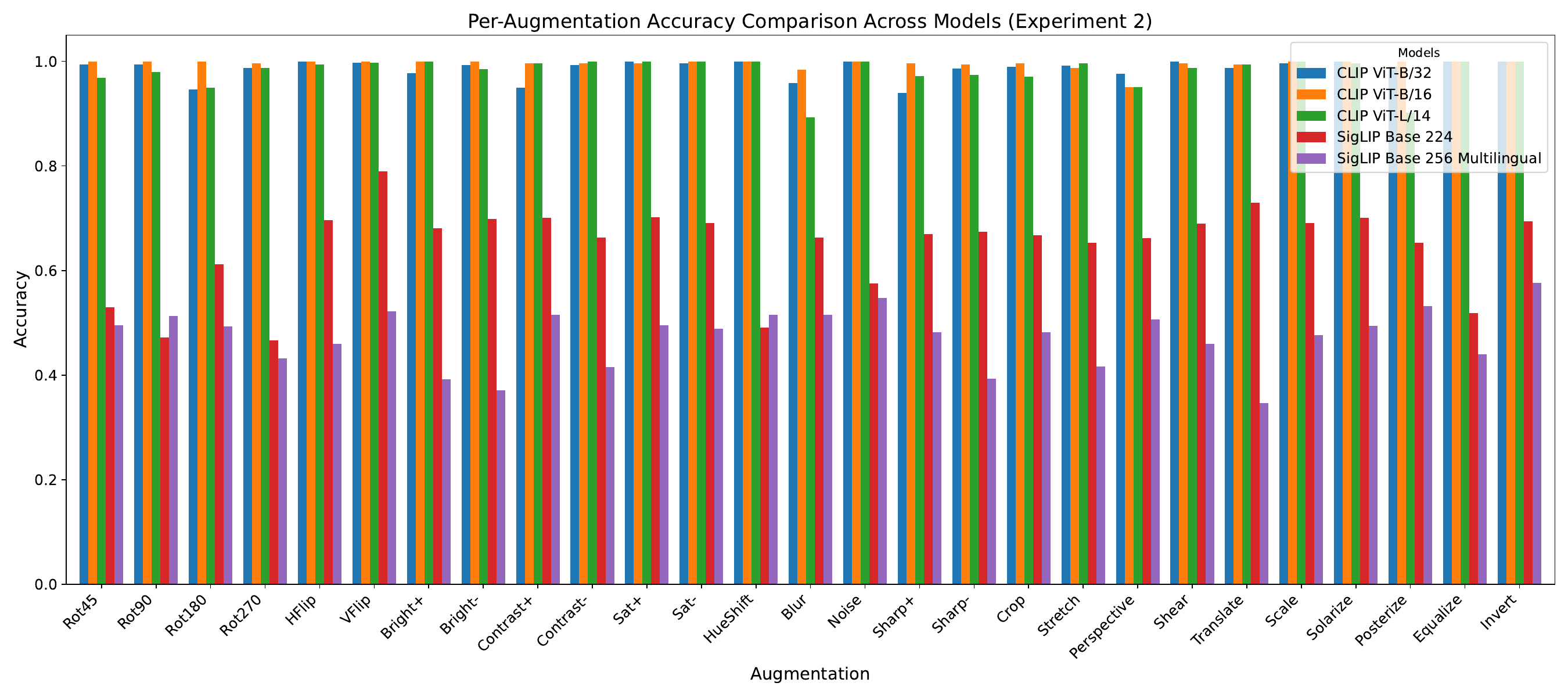}
    \caption{Per Augmentation Accuracy Experiment 2}
    \label{fig:exp2_results_per_aug}
\end{figure*}

\begin{figure}[t]
    \centering
    \includegraphics[width=0.9\linewidth]{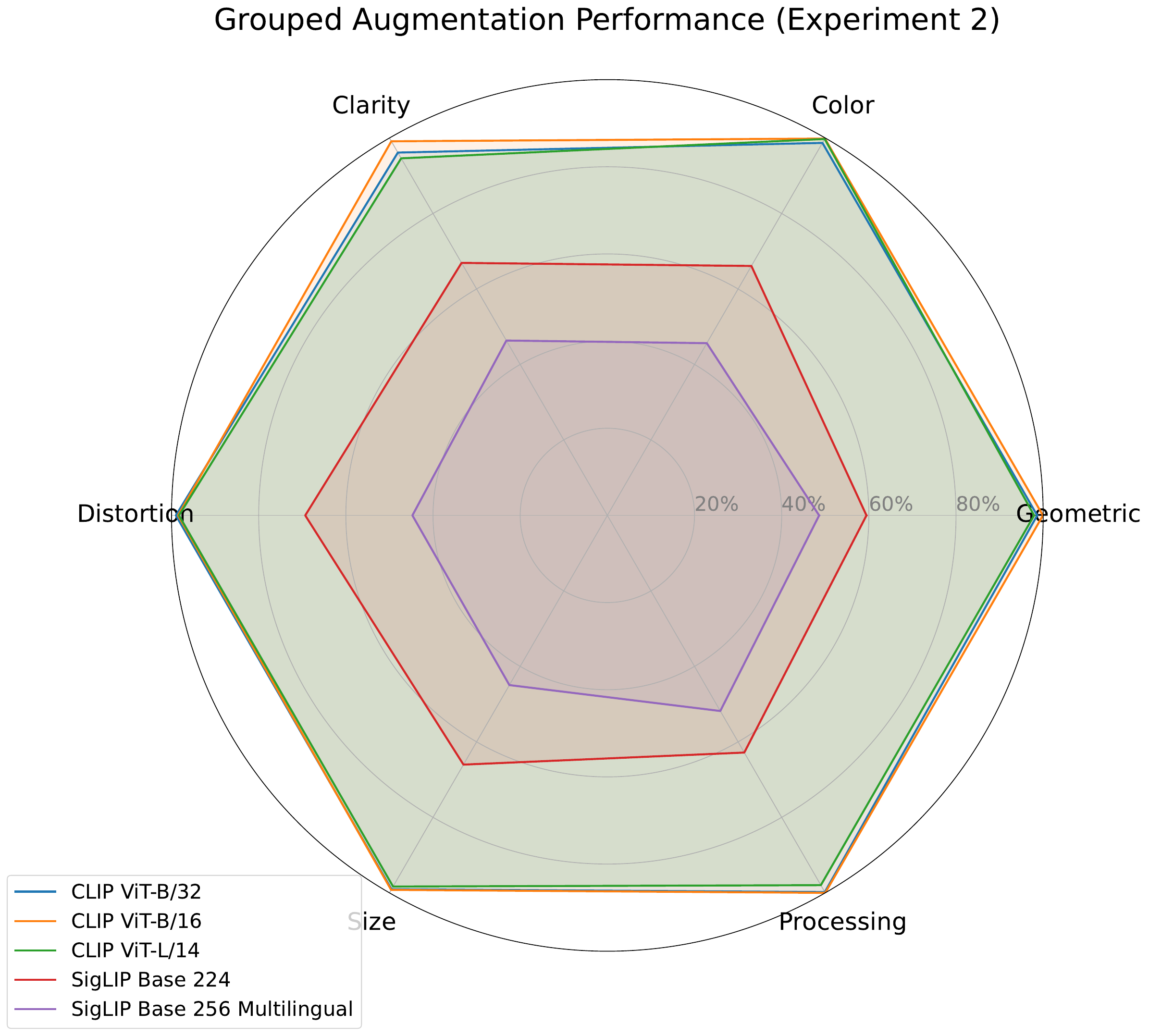}
    \caption{Per Augmentation Accuracy Experiment 2}
    \label{fig:exp2_results_radio}
\end{figure}

Figure \ref{fig:exp2_results_per_aug} shows the results of CLIP and SigLIP for experiment 2 per augmentation category, these results reflect our initial analysis that CLIP model is performing well in differentiating between original prompt and augmented prompt when we calculate the similarity.
Figure \ref{fig:exp2_results_radio} shows the performance of CLIP and SigLIP when grouped by the categories mentioned earlier and further strengthen the result that CLIP models are performing better.

\subsection{Classifying Image Transformations}
This evaluation assesses the ability of VLMs to accurately identify specific image transformations from a predefined set of augmentations. Unlike the previous evaluations, which focused on pairwise comparisons, this assessment tests the models' direct classification capabilities across a comprehensive range of augmentation types. By examining how well the models can classify various image modifications, we can better understand their ability to recognize and categorize different types of visual changes.

\subsubsection{Methodology}

% \begin{figure}[t]
%     \centering
%     \includegraphics[width=0.9\linewidth]{fig/category_accuracy_radar_chart_exp3.pdf}
%     \caption{Accuarcy of Models in understanding augmentation}
%     \label{fig:exp3_radar_chart}
% \end{figure}

\begin{figure*}[h!]
    \centering
    \includegraphics[width=0.9\linewidth,height=7cm]{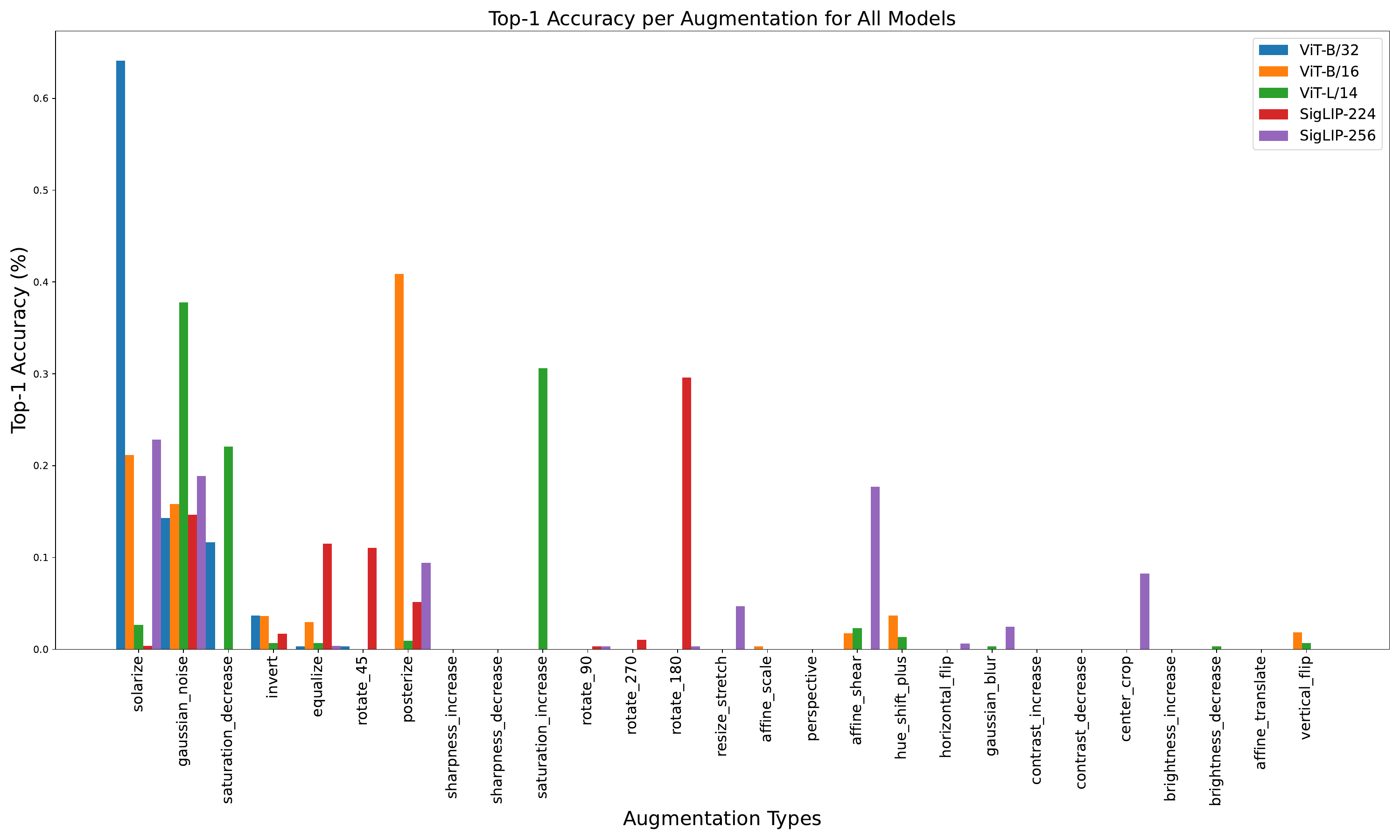}
    \caption{Top-1 Accuracy per Augmentation type for all models}
    \label{fig:exp3_results_acc}
\end{figure*}

\begin{table*}[ht]
    \centering
    \caption{Comparison of Top-1 and Top-5 Accuracies for Each Model}
    \label{tab:accuracy_comparison}
    \begin{tabular}{@{}lcc@{}}
        \toprule
        \textbf{Model} & \textbf{Top-1 Accuracy (\%)} & \textbf{Top-5 Accuracy (\%)} \\
        \midrule
        ViT-B/32 & 3.61 & 18.40 \\
        ViT-B/16 & 3.50 & 17.12 \\
        ViT-L/14 & 3.57 & 15.28 \\
        SigLIP Base 224 & 2.81 & 16.40 \\
        SigLIP Base 256 Multilingual & 3.19 & 18.06 \\
        \bottomrule
    \end{tabular}
\end{table*}

For each augmented image \( I_{\text{aug}} \), we perform the following steps:

First, we present the model with the augmented image \( I_{\text{aug}} \) and compare it against all possible augmentation descriptions \( \mathcal{A} \), consisting of 27 types as described in Section \ref{subsec:datacollection}. For each augmentation description \( a \in \mathcal{A} \), we calculate the similarity score between the image and the textual description:
\begin{equation}
    \text{score}_a = \text{sim}\left( I_{\text{aug}},\, \text{``} a \text{''} \right)
\end{equation}

We rank all augmentation descriptions based on their similarity scores in descending order. The rank of the true augmentation description \( t \) is determined by:
\begin{equation}
    \text{rank}_t = \left| \left\{ a \in \mathcal{A} : \text{score}_a > \text{score}_t \right\} \right| + 1
\end{equation}

We evaluate the model's performance using the following metrics:
\begin{itemize}
    \item \textbf{Top-1 Accuracy}: The proportion of times the correct augmentation \( t \) is ranked first (\( \text{rank}_t = 1 \)).
    \item \textbf{Top-5 Accuracy}: The proportion of times \( t \) is among the top five predictions (\( \text{rank}_t \leq 5 \)).
    \item \textbf{Mean Rank}: The average rank position of the correct augmentation \( t \) across all samples.
\end{itemize}

This approach assesses the model's ability to accurately identify the augmentation applied to an image by matching it with the correct textual description.

\subsubsection{Results}
This experiment shows Vision Language Understanding of Augmentations where can a model associate itself with the correct Augmentation. Figure \ref{fig:exp3_results_acc} shows the Top-1\% accuracy performance of Vision Language Models on just identifying the correct augmentation class where for most of the augmentation, the accuracy is 0\% and model was not able to identify a single correct example. Table \ref{tab:accuracy_comparison} compares the Top-1\% and Top-5\% accuracy and shows that Vision Language Model can not classify the augmentation correctly. 
% Figure \ref{fig:exp3_radar_chart} shows that while Vision Language Models are bad at classifying the correct augmentation, they perform extremely poor at processing augmentations such as translation, rotation, scaling, and cropping, these are the transformations where humans excel in general.
\section{Impact on Downstream task}
With the rise of AI in Image/Video Editing\cite{tang2024exploringimpactaigeneratedimage}, this study reveals an important lack of understanding of the image level in vision language models. These models, predominantly built on CLIP\cite{radford2021learningtransferablevisualmodels} as their backbone architecture, form the foundation of numerous downstream tasks such as Image Generation\cite{rombach2022highresolutionimagesynthesislatent}\cite{wang2021actionclipnewparadigmvideo}, Controlled Image Generation models \cite{zhang2023addingconditionalcontroltexttoimage}\cite{li2024controlnetimprovingconditionalcontrols}\cite{zavadski2024controlnetxsrethinkingcontroltexttoimage}\cite{qin2023unicontrolunifieddiffusionmodel}, Image-to-Image Editing\cite{parmar2023zeroshotimagetoimagetranslation}\cite{choi2023customedittextguidedimageediting}\cite{matsunaga2023finegrainedimageeditingpixelwise} and multiple other downstream tasks.
CLIP-based architectures, which align visual and textual representations through contrastive learning, have demonstrated remarkable capabilities in understanding semantic content. However, our analysis exposes a critical limitation in their spatial understanding of images. Different types of image transformation are a basic tool in traditional image editing tools such as Photoshop\cite{adobe_transform_2024}, yet modern AI systems struggle with these operations. Table \ref{tab:qual_analysis} shows examples of common AI Image editing models, Instruct Pix2Pix\cite{brooks2023instructpix2pixlearningfollowimage}, Dall.E 3\cite{dalle3}, and IP Adapter\cite{ye2023ipadaptertextcompatibleimage} with the prompt \textbf{"Rotate the input image 90 degrees"}. The results demonstrate that none of these CLIP-powered models was able to understand this basic instruction and failed to generate an image with the requested transformation applied.
This fundamental limitation suggests that despite their impressive semantic capabilities, current CLIP-based models lack a comprehensive understanding of image structure and spatial relationships due to their invariant nature which comes at the cost of explicit spatial understanding. This paper motivates us to think about newer training paradigms for Vision Language Models that balance invariance with explicit transformation awareness, where models can have global context, understand images at a deeper structural level beyond just semantic content, and reason about spatial manipulations when required. Addressing these limitations will help unlock newer capabilities in downstream tasks, potentially bridging the gap between AI-powered and traditional image editing tools.

% \vspace{-10pt}  % Negative space before table
\begin{table*}[!htb]
\centering
\renewcommand{\arraystretch}{1.5} % Adjust row height
\begin{tabular}{|c|c|c|}
\hline
\textbf{Model} & \textbf{Input Image} & \textbf{Output Image} \\ \hline
DALL·E & 
\includegraphics[width=0.2\textwidth]{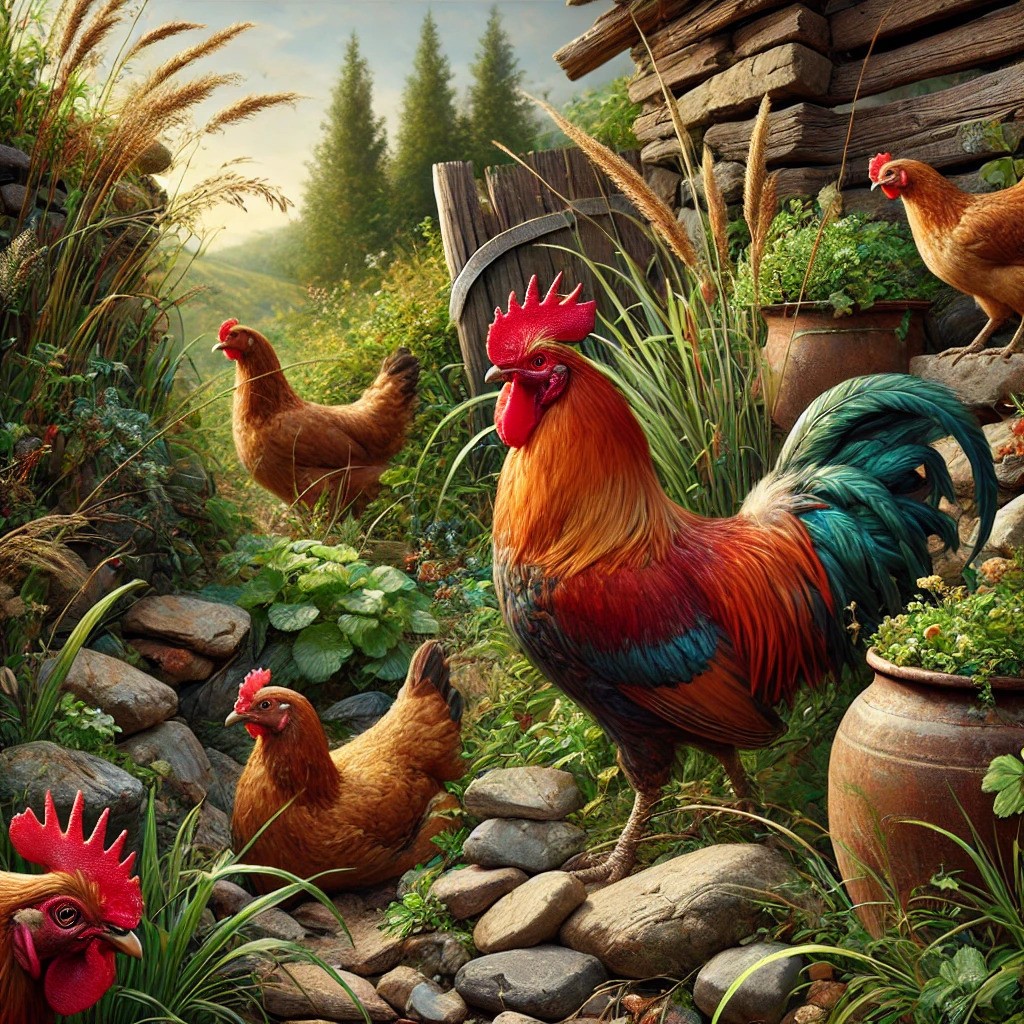} & 
\includegraphics[width=0.2\textwidth]{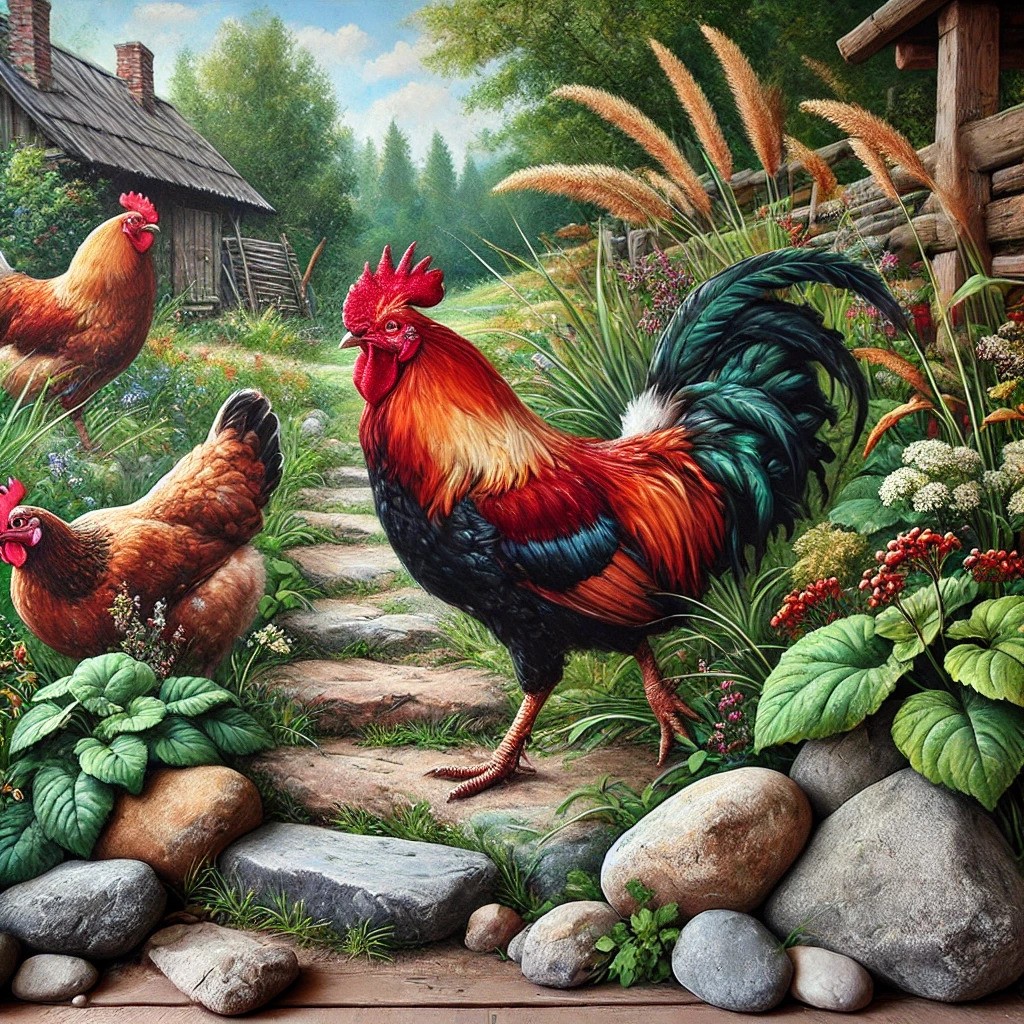} \\ \hline
Instruct Pix2Pix &
\includegraphics[width=0.2\textwidth]{fig/14866578404_d4ba6f82be_c.jpg} & 
\includegraphics[width=0.2\textwidth]{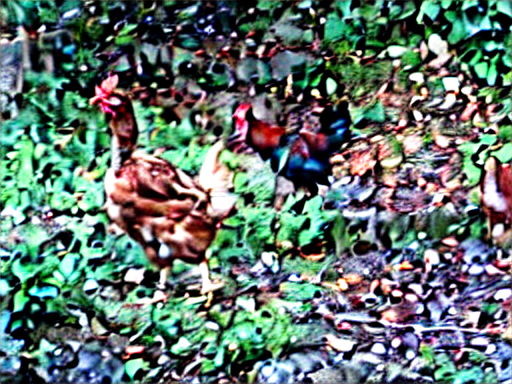} \\ \hline
IP Adapter & 
\includegraphics[width=0.2\textwidth]{fig/14866578404_d4ba6f82be_c.jpg} & 
\includegraphics[width=0.2\textwidth]{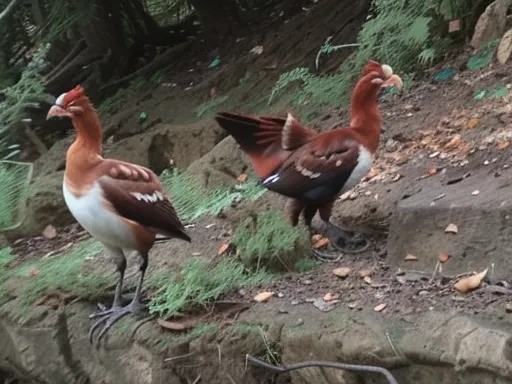} \\ \hline
\end{tabular}
\caption{Qualitative analysis table comparing input images and output transformations (rotation 90 degrees) for different models.}
\label{tab:qual_analysis}
\end{table*}

% \section{Acknowledgement}
% Valuable comments and insightful feedback from Anas Zafar are gratefully acknowledged.

\clearpage
{
    \small
    \bibliographystyle{ieenat_fullname}
    \bibliography{main}
}
\end{document}